# JRC EuroVoc Indexer JEX – A freely available multi-label categorisation tool

**Ralf Steinberger,  Mohamed Ebrahim  &  Marco Turchi**

European Commission – Joint Research Centre (JRC)

Via Fermi 2749, 21027 Ispra (VA), Italy

E-mail: Firstname.Lastname@jrc.ec.europa.eu

**Abstract**

EuroVoc (2012) is a highly multilingual thesaurus consisting of over 6,700 hierarchically organised subject domains used by European Institutions and many authorities in Member States of the European Union (EU) for the classification and retrieval of official documents. JEX is JRC-developed multi-label classification software that learns from manually labelled data to automatically assign EuroVoc descriptors to new documents in a profile-based category-ranking task. The JEX release consists of trained classifiers for 22 official EU languages, of parallel training data in the same languages, of an interface that allows viewing and amending the assignment results, and of a module that allows users to re-train the tool on their own document collections. JEX allows advanced users to change the document representation so as to possibly improve the categorisation result through linguistic pre-processing. JEX can be used as a tool for interactive EuroVoc descriptor assignment to increase speed and consistency of the human categorisation process, or it can be used fully automatically. The output of JEX is a language-independent EuroVoc feature vector lending itself also as input to various other Language Technology tasks, including cross-lingual clustering and classification, cross-lingual plagiarism detection, sentence selection and ranking, and more.

## 1.  Introduction

Many parliaments and other national and international authorities employ librarians that manually assign EuroVoc descriptors (subject domain labels) to their documents. EuroVoc (2012) is a wide-coverage multidisciplinary thesaurus with over 6,700 classes covering the activities of the EU, and in particular those of the European Parliament. The EuroVoc labels have been translated one-to-one into at least 27 languages.[1] Due to the large number of categories, the human classification process is very complex and thus slow and expensive: Human documentalists index between thirty and thirty-five documents per day. While the automatic assignment performs less well than the manual assignment, it has the advantage that it is fast and consistent. An *interactive* usage of the tool hopefully helps improve both speed and consistency of the human annotation process. However, the tool also allows to fully automatically index large document collections that would otherwise never be indexed manually, due to the high cost and the non-availability of human resources.

Apart from being a support tool for the human indexers, the language-independent output of the multi-label classification software can also be used as input to other text mining applications, e.g. for the detection of document translations or plagiarised text (Pouliquen et al. 2004; see also Potthast et al. 2010 and the references therein); to link related documents across languages (Pouliquen et al. 2008); to support the lexical choice in Machine Translation; to rank sentences in topic-specific summarisation, and more.

The EuroVoc thesaurus is managed and maintained by the European Union's *Publications Office* (PO), which moved forward to ontology-based thesaurus management and semantic web technologies conformant to W3C recommendations, as well as to latest trends in thesaurus standards. EuroVoc users include the European Parliament, the Publications Office, national and regional parliaments in Europe, plus national governments and other users around the world.

The Spanish *Congress of Deputies* in Madrid has been using the JRC-developed software in their daily indexing workflow since 2006. Since further parliaments expressed their interest in using the software, the JRC re-implemented and distributed it under the name of JEX. The decision to release JEX freely to the wider public is to be seen in the context of Directive 2003/98/EC of the European Parliament and of the Council on the re-use of public sector information.[2] This directive recognises that public sector information such as multilingual collections of documents can be an important primary material for digital content products and services, that their release may have an impact on cross-border exploitation of information, and that it may thus have a positive effect on an unhindered competition in the EU's internal market. For the same reason, the JRC and its partners have distributed a number of further highly multilingual resources publicly, including the *JRC-Acquis* parallel corpus (Steinberger et al. 2006), the *DGT-TM* translation memory (Steinberger et al. 2012) and the named entity resource *JRC-Names* (Steinberger et al. 2011).

---

[1] EuroVoc exists not only in 22 official EU languages (Bulgarian, Czech, Danish, Dutch, English, Estonian, Finnish, French, German, Greek, Hungarian, Italian, Latvian, Lithuanian, Maltese, Polish, Portuguese, Romanian, Slovak, Slovenian, Spanish and Swedish), but also in Basque, Catalan, Croatian, Russian and Serbian. Further non-official translations exist.

[2] For details and to read the full text of the regulation, see http://eur-lex.europa.eu/LexUriServ/LexUriServ.do?uri=CELEX:32003L0098:EN:NOT. All URLs were last visited in March 2012.

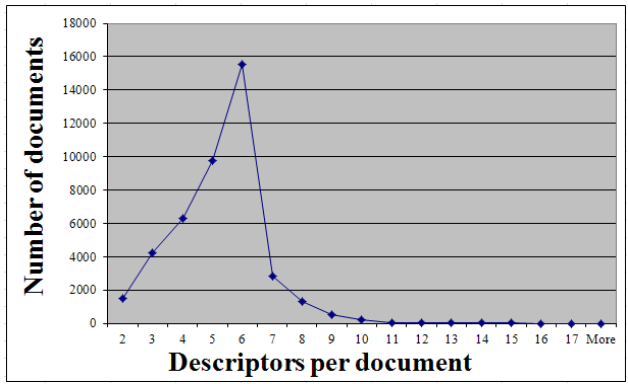

**Fig. 1.** Distribution of EuroVoc classes per document.

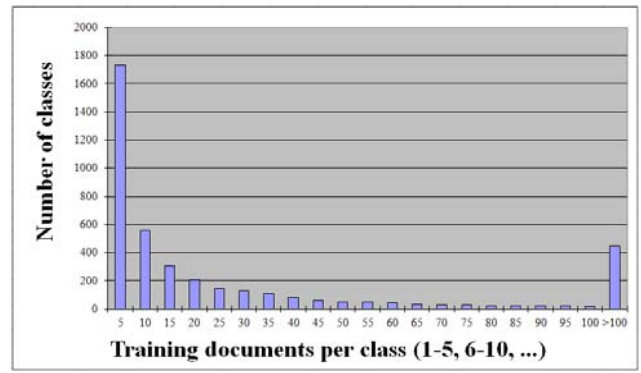

**Fig. 2.** Imbalanced usage of EuroVoc descriptors.

The remainder of this paper is organised as follows: We first discuss related work (Section 2) and describe the document collection on which JEX was trained and tested (3). Then we give an overview of our own categorisation algorithm (4) and we discuss its performance (5). Section 6 describes the functionality of JEX and provides some practical usage information.

## 2. Related work

Automatic EuroVoc indexing is a multi-label classification task, discussed abundantly in literature (e.g. Tsoumakas and Katakis, 2007). Here, we focus specifically on EuroVoc-related work: Loza Mencia & Fürnkranz (2010) applied three different multi-label classification algorithms to the (EuroVoc-indexed) Eur-Lex database of the European Union addressing the problem of storing in memory all the trained classifiers. Using the Dempster-Shafer Theory, Sarinnapakorn & Kubat (2007) combined the output of several sub-classifiers to determine which EuroVoc thesaurus classes to assign to documents. The Text Categorization Toolkit TECAT (Montejo Raez, 2006) was used to test different classifiers on the EuroVoc collection. All of the authors mentioned applied several machine learning techniques that mstly rely on positive and negative samples for each class. De Campos and Romero (2009) automatically associated an ordered set of appropriate descriptors extracted from EuroVoc to a document to be classified creating a unique Bayesian network for the full thesaurus; this approach can be used with or without training data.

Some authors investigated whether changing the document representation improves the indexing performance: Daudaravičius (2012), working with English, Finnish and Lithuanian, showed that EuroVoc indexing performance can be improved significantly when applying collocation segmentation, and he observed that the indexing performance for longer documents was better than for shorter ones. Ebrahim et al. (2012) verified whether lemmatisation and part-of-speech tagging would improve the indexing performance for languages from different language families (Czech, English, Estonian and French, covering a Slavic, a Germanic, a Finno-Ugric and a Romance language). They came to the conclusion that lemmatisation did not improve the performance for any of the languages, but that part-of-speech-tagging mostly improved the results slightly. In their experiments, results using the mean mutual rank always yielded less good results, while using the merged average was promising.

The EuroVoc thesaurus and all the legal documents indexed with its categories have been widely used to generate resources in natural language processing. Erjavec et al. (2005) compiled a massively multilingual corpus, the *EU Acquis*, introducing the corpus annotation tool *totale*. Steinberger et al. (2006) created a widely used multilingual aligned parallel corpus, the *JRC-Acquis*, from manually labelled documents in 22 languages.

In the approach presented here, we focus only on the positive available training points to extract a representative vector for each category (the *category profile*), using single inflected words. The choice of using only positive samples speeds up the training phase, reduces the computational cost of storing in memory large numbers of classifiers and avoids the problem of unbalanced positive and negative data. We distribute the training data used with the JEX release, together with information on the exact split of the training and test collection into ten parts (for ten-fold cross-validation) in order to allow best possible comparison with other systems.

## 3. Details on the document collection used

The number of assigned EuroVoc categories per document in this collection ranges from 2 to 17, with the distribution shown in **Figure 1**, an average of 5.43 descriptors per document and a standard deviation of 1.86. **Figure 2** shows that the training collection is highly imbalanced: 1733 descriptors were used five times or less. 355 descriptors were used between fifty-one and one hundred times. 420 descriptors were used between one hundred and one thousand times, and 27 descriptors more than one thousand times. The most frequently used descriptor was used 4262 times. This high number of classes, combined with a very unevenly balanced training set, is a big challenge for a multi-label categorisation task. Note that, with the JEX parameter settings used in the experiments described in this chapter, we require a minimum of four documents to train a category.

## 4. EuroVoc indexing methodology

This paper is mostly about the public release of the software and its training data. For details on the classification algorithm, see Pouliquen et al. (2003).

The **major challenges** for EuroVoc indexing are (a) the high number of different categories (thousands), (b) the multi-label training data (between two and seventeen categories per text), and (c) the highly imbalanced distribution of these different categories. We address the challenge by treating it like a **profile-based category ranking task**. The profile of each class (we refer to *classes* also as *descriptors* or *categories*) consists of a ranked list of typical features for this class. In the experiments described here, features are inflected word tokens, but the software also allows using lemmas, word n-grams, segments, word_part-of-speech combinations, and more. We refer to these features as *associates*. The document to be indexed is represented as a vector of the same features (inflected word forms, n-grams, etc.) with their frequency in the document. The training documents, on the other hand, are represented as a log-likelihood-weighted list of features, using the training document set as the reference corpus.[3] The most appropriate categories for the new document are found by ranking the category vector representations (the *profiles*) according to their cosine similarity with the vector representation of the document to be indexed. Our major effort was thus spent on optimising the profile of each class. JEX uses large numbers of stop words that are ignored in the classification process. Improving the stop word lists was identified as the most efficient way of improving the indexing performance: Experiments on Spanish presented in Pouliquen et al. (2003) yielded an improvement of 5.6 points on the F1 scale when using a good stop word list, while lemmatisation and the usage of multi-word term lists improved the results by 2 and 0.1 points, respectively.

In order to optimise the profile generation for each class, a number of different parameter settings were optimised by selecting the best-performing setting within a range of values. The following are some of the most important (and easiest to explain) parameters used (see Pouliquen et al. 2003 for more details). The numbers in brackets show the settings used in the experiments described in this article:

1. How many training documents there must be at least for a class to be trained (4);
2. how long these training documents must be at least (100 words);[4]
3. how often words need to be found in the corpus in order to be used as associates (4);
4. how statistically relevant a word must be in a training document in order to be considered (minimum log-likelihood value of 5);
5. how to weight words depending on the number of descriptors assigned to each training document.

The parameters used in our experiments were experimentally optimised in the past using an English training set. Ideally, the parameters should be optimised for each document representation and for each language, but we do not expect this to make a big difference.

Using word bigrams instead of single words also improved the optimal F1-score in our experiments by about one point, but the word bigrams are not used in the JEX default parameter settings, for two reasons: bigram usage leads to increased vocabulary dimension and sparseness; also, we wanted to achieve simplicity of usage and transparency towards the librarian end user when displaying the associates.

## 5. Evaluation / Discussion of results

The system assigns a ranked list of descriptors rather than a set. As there is an average of just under six descriptors per document in the training set, we evaluated the performance for the top six automatically assigned descriptors (rank 6), independently of how many descriptors were actually assigned manually. When more descriptors had been manually assigned, it was thus impossible to achieve 100% Recall; when less descriptors had been assigned manually, the system could not achieve 100% Precision. We additionally evaluated the performance using a *dynamic rank*, meaning that we evaluated the performance for each document at rank X, where X is the number of descriptors that had been assigned manually to that document. Note that – while using evaluation at dynamic rank yields better evaluation results for all languages – it is not realistic because in a real-life setting the documents will not yet have been indexed manually and the dynamic rank will be unknown.

We automatically evaluated the performance of the system for each of the languages through ten-fold cross-validation, using a collection of tens of thousands of manually labelled Eur-Lex documents (European legislation). Results are shown in **Table 1**.[5] *Recall* is defined as the number of correct and automatically assigned descriptors divided by the number of all manually assigned categories. *Precision* is defined as the number of correct and automatically assigned descriptors divided by the total number of automatically assigned descriptors (in this case always six, except in the column *dynamic rank*).

Generally speaking, we note that the difference in performance (F1) between the twenty-two languages is rather comparable (except for Maltese), ranging from F1=0.48 for Romanian to 0.53 for Hungarian. This is an encouraging result as the number of training documents differs a lot between the different languages, and as these 22 languages are from very diverse language families: Germanic, Romance, Slavic, Hellenic, Finno-Ugric, Baltic and Semitic.

The column 'Total number of documents' in Table 1

---

[3] In the work described in Pouliquen et al. (2003), we experimented with various other representations (e.g. pure frequency and TF.IDF), but the one adopted performed best.

[4] Results presented in Daudaravičius (2012) indicate that this parameter should be tuned for the various languages.

[5] Note that the results reported in Pouliquen et al. (2003) were achieved using more documents, but consisting of a mix of different types.

| Language | Precision | Recall | F1 | F1 dynamic rank | Categories / collection | Average categories trained | Stop words used (+MW) | Total number of documents | Document length (words) ± std-dev | All categories trained |
|---|---|---|---|---|---|---|---|---|---|---|
| BG | 0.4619 | 0.5120 | 0.4857 | 0.4940 | 3780 | 2049.9 | 332 | 22696 | 786.96±2784.72 | 2147 |
| CS | 0.4689 | 0.5205 | 0.4933 | 0.4990 | 3691 | 2035.7 | 137 | 22830 | 890.66±3317.10 | 2129 |
| DA | 0.4747 | 0.5491 | 0.5092 | 0.5170 | 4226 | 2655 | 858 | 41727 | 561.87±1875.19 | 2752 |
| DE | 0.4732 | 0.5485 | 0.5081 | 0.5187 | 4230 | **2683.9** | 1793 | 41676 | 566.67±1945.44 | **2783** |
| EL | 0.4632 | 0.5369 | 0.4973 | 0.5118 | 4214 | 2486.4 | 105 | 41103 | 778.05±2379.45 | 2605 |
| EN | 0.4801 | 0.5547 | 0.5147 | 0.5227 | 4229 | 2324.1 | 1972 (+545) | 41672 | 309.28±1176.39 | 2430 |
| ES | 0.4801 | 0.5545 | 0.5147 | 0.5188 | 4221 | 2297 | 481 (+264) | 41397 | 547.63±1819.14 | 2406 |
| ET | 0.4828 | 0.5358 | 0.5079 | 0.5139 | 3662 | 2047.8 | 1533 | 21989 | 652.22±2193.32 | 2147 |
| FI | 0.4654 | 0.5341 | 0.4974 | 0.5081 | 4103 | 2528.8 | 92 | 38293 | 756.70±2565.81 | 2634 |
| FR | 0.4776 | 0.5536 | 0.5128 | 0.5227 | **4234** | 2588.7 | 1180 | **41989** | 663.33±2204.28 | 2688 |
| HU | **0.5121** | **0.5654** | **0.5374** | **0.5444** | 3585 | **1688.5** | 1228 (+709) | 20838 | 551.66±1977.14 | **1788** |
| IT | 0.4713 | 0.5464 | 0.5061 | 0.5151 | **4234** | 2584.4 | 219 | 41838 | 764.57±2808.31 | 2688 |
| LT | 0.4920 | 0.5454 | 0.5174 | 0.5239 | 3635 | 1945.7 | 1199 | 21505 | 644.53±2724.18 | 2046 |
| LV | 0.4659 | 0.5175 | 0.4904 | 0.4968 | 3690 | 2011 | 14 | 22803 | 894.59±3012.39 | 2106 |
| MT | **0.4200** | **0.4545** | **0.4366** | **0.4416** | **3584** | 1762.3 | **6** | 17858 | 1016.99±2685.11 | 1864 |
| NL | 0.4803 | 0.5562 | 0.5155 | 0.5257 | 4232 | 2610.2 | 1414 | 41816 | 581.94±1819.66 | 2713 |
| PL | 0.4794 | 0.5311 | 0.5039 | 0.5077 | 3648 | 1967.1 | 125 | 22004 | 841.81±2795.35 | 2066 |
| PT | 0.4756 | 0.5493 | 0.5098 | 0.5237 | 4209 | 2560.6 | 1152 | 41142 | 700.46±2138.01 | 2663 |
| RO | 0.4550 | 0.5043 | 0.4784 | 0.4817 | 3887 | 2109.3 | 1504 (+48) | 25023 | 994.17±3083.16 | 2206 |
| SK | 0.4705 | 0.5204 | 0.4942 | 0.4995 | 3645 | 1938.4 | 364 | 21406 | 872.53±3241.50 | 2050 |
| SL | 0.4840 | 0.5341 | 0.5078 | 0.5205 | 3685 | 2013.1 | 2068 | 22289 | 627.56±2669.38 | 2119 |
| SV | 0.4787 | 0.5473 | 0.5107 | 0.5194 | 4109 | 2546.4 | 1093 | 38198 | 609.82±2365.26 | 2649 |

**Table 1.** Statistics on the training document set and on the evaluation results of the EuroVoc descriptor assignment for 22 languages. Highest and lowest numbers for each relevant column are highlighted in bold-face. The evaluation results in columns 2 to 7 are based on ten-fold cross-validation. The last four columns describe the whole document collection (combined training and test set). The last column indicates how many EuroVoc descriptor classifiers have been trained (using *all* documents of the collection; no testing) in the software released publicly.

shows that, for half of the 22 languages, our document collection consists of around 41,000 training and test documents per language, and for the other half it consists of about 22,000. The languages with less documents are those that joined the EU more recently so that the body of EU law is smaller for these languages. As the countries joining the EU more recently had to translate and adopt all the currently valid EU legislation, we can conclude that the documents that were *not* translated (and used for our training) are the older ones that are no longer valid legislation. Presumably, the vocabulary of the older documents is also different and it might thus be a good idea to ignore these documents also for the languages that have about 41,000 training documents. The correlation between F1 and the **document collection size** (according to Table 1) is 0.38. A scatter plot for these two columns shows rather clearly that the languages with many training documents produce a stable F1 performance of between 0.50 and 0.52, while the ones with a smaller document collection produce rather heterogeneous results (F1 varying between 0.44 and 0.54).

The column 'Categories/collection' shows how many distinct categories are present in the document collection, and 'Average categories trained' shows how many category classifiers were trained on average in the ten-fold cross-validation experiments. 'All categories trained' indicates the number of classifiers trained in the final software release, where 100% of the document collection was used for training. Note that any classification task becomes significantly more difficult if more categories need to be assigned, meaning that the **number of categories in the collection** must be considered when analysing the classification performance (P, R and F1). It is probably not coincidental that Hungarian, which performed best of all languages (Hungarian, F1=0.54), is also the language for which least classifiers could be trained (1688). The correlation values between F1 and categories/collection and average categories trained are 0.31 and 0.29, respectively.

Maltese performed least well of all languages

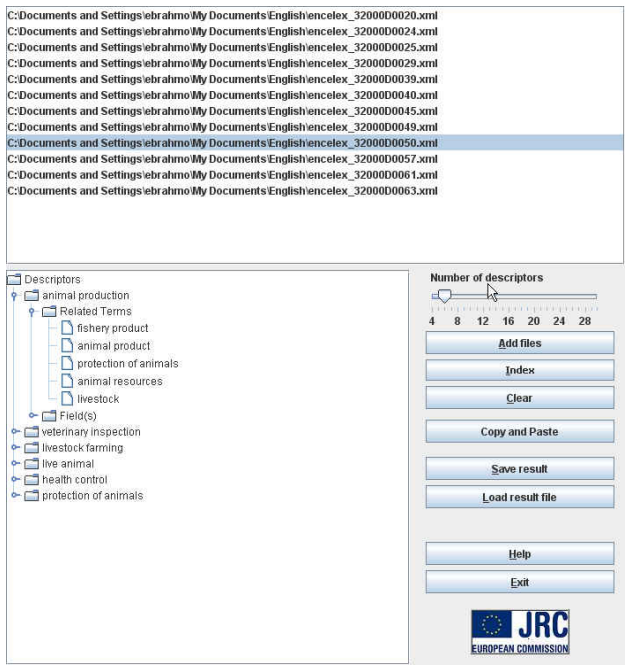

**Figure 3.** GUI of the JEX tool, showing documents to be indexed (top window), automatically assigned descriptors plus their hierarchical position inside EuroVoc (left) and various options (right).

(F1=0.44), about four points below the second worst-performing language Romanian. This low performance could be related to the fact that Maltese has the smallest document collection (17,858 documents), but it could also be related to the extremely small size of the Maltese stop word list (six elements). In our experiments, we identified a good **stop word list** as the easiest and most efficient way of improving classification performance. For some of the languages, we also used multi-word (MW) stop words, i.e. groups of adjacent words that were ignored during the classification process. Multi-word stop words allow to suppress highly repetitive phrases, which are very common in legal text, such as *having regard to Commission Decision Number*. The assignment algorithm does of course suppress the assignment impact of highly frequent words, but as both descriptors and simple words like *question* and *authorities* may be distributed unevenly across different text types, there is always a chance that these words get some undeserved weight in the assignment process. In our experience, a good list of stop words hand-selected by persons who know both EuroVoc and the document collection is very beneficial for the system performance. Our observation is confirmed by the high correlation (0.53) between F1 and the size of the stop word list in Table 1. This value is significantly higher than all other correlation values.

When looking at the relationship between performance and **language family**, some tendencies can be observed: All five Germanic languages (Nl, En, Sv, Da, De) are among the ten best-performing, while all five Slavic languages (Sl, Pl, Sk, Cs, Bg) are in the bottom half. The tendencies are much less clear for the remaining language families: Three out of five Romance languages (Es, Fr, Pt) are in the better half, while the other two are not (It, Ro); The Finno-Ugric languages (Hu, Et, Fi) are first, eleventh and fifteenth; The Baltic languages (Lt, Lv) are in second and thirteenth position; Maltese (Mt; last position) is the only Semitic language so that generalising is not useful.

In spite of all the differences mentioned above, the overall comparable results across languages (F1=0.48 to 0.54) indicate that the algorithm and the default parameter settings are seemingly nearly language-independent. This is not obvious, as the ratio of word forms per lemma is much higher for highly inflected languages than for lesser inflected ones, meaning that there is less lexical evidence in highly inflected languages and one might thus assume that relatively more training material is needed. At the same time, the experiments described by Ebrahim et al. (2012) indicate that reducing the type-token ratio through lemmatisation was consistently counter-productive when applied to one representative each of the Germanic, Romance, Finno-Ugric and Slavic language families.

Remains the question what could be done to significantly improve the classification results. Is it possible to achieve F1 values of 0.80 or more? A manual evaluation of automatically assigned descriptors carried out for English and Spanish showed that the top 5 automatically assigned descriptors were perfectly adequate (75% and 82% were judged to be good, respectively) even if they had *not* initially been assigned manually (Pouliquen et al. 2003). The same experiment furthermore showed that the second annotator agreed with the first annotator in 74% (English) and 84% of cases (Spanish). These human agreement values can thus be seen as an upper bound performance for this highly complex and subjective multi-label classification task involving thousands of categories. We nevertheless hope that making JEX and the document collection available to the research community will encourage further research and that this will eventually result in better-performing tools.

## 6. Downloading and using JEX

The readily trained software is distributed under the European Union Public Licence (EUPL) and it can be downloaded from the JRC's Language Technology site http://langtech.jrc.ec.europa.eu/JRC_Resources.html. It provides a graphical user interface (GUI), a command line interface and an API. This gives users the freedom to configure the system and to adapt it to their own needs.

The **GUI** provides the librarian end user with an easy and intuitive way of using the system interactively. Users can either cut-and-paste a single text to be EuroVoc-indexed, or they can load a set of documents by browsing the file system. JEX accepts files in plain text, XML or HTML. The contents of tags will be ignored; all other text will be used in the assignment process. Users can select how many automatically assigned descriptors they want to see and then launch the indexing process (**Figure 3**). They can view the results, document by document, and they can choose to view the hierarchical position of each descriptor inside the EuroVoc hierarchy: broader terms, narrower terms, related terms and fields (in EuroVoc, *fields* are higher-level containers for categories). By opening an individual document, the users can see the full text (**Figure 4**) and – by selecting any of the assigned Eurovoc descriptors, they see a list of all associates of this descriptor, as well as the associates found in the document highlighted in the text. This latter functionality allows the user to quickly judge why the assigned descriptor was assigned. Should users disagree with the automatic assignment, they can delete an automatically assigned descriptor or they can add a new descriptor manually by searching or browsing the EuroVoc hierarchy. The results can then be saved into an XML file, as shown in **Figure 5**. The output consists of the document identifier followed by the list of automatically assigned descriptors and their respective weight (assignment strength). If the input file is an XML file, the output will be added directly to the input XML file.

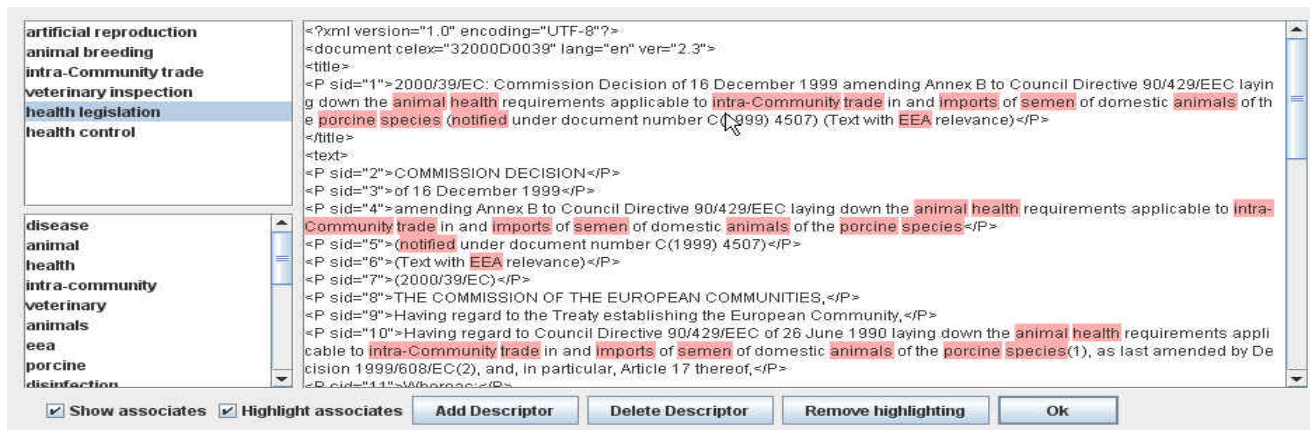

**Figure 4.** JEX GUI with automatically assigned EuroVoc desriptors (top left) for a document (right) and - for the selected descriptor - the list of its associates (bottom left). The associates found are also highlighted in the document (right).

We also provide command-line tools for Windows and un\*x-like systems. The **command line interface** allows batch-indexing a set of documents, re-training the system on a new set of documents, and automatically evaluating the performance of the tool.

Indexing a set of documents using the command line interface is as easy as putting these documents in a directory and telling the system through a configuration file where this directory is.

Users can configure various parameters and settings that may have an impact on the indexing performance (e.g. those discussed in Section 4) or that determine the output format of the tool, but doing this is not obligatory because the default settings should mostly work well. Users can furthermore make changes to the stop word lists and they can include a blacklist of descriptors that should never be assigned. This blacklist may be useful, for instance, to exclude EuroVoc descriptors that were assigned to documents in the past, but that are no longer used.

The system also allows users to **retrain its classifiers** for various purposes: (a) to improve the indexing performance by adding more training documents; (b) to additionally train it on user-specific descriptors; (c) to train the system on a document collection indexed with an entirely different thesaurus; (d) to test various parameter settings in order to identify the best-performing ones; (e) to experiment whether other document representations produce better results (e.g. lemmatised word forms, word n-gram representations, etc.). Retraining the system does not involve much effort once the training document set is in the input format required by JEX, called the *compact format*. Compact format files are text files containing all training documents. For each training document, the first line contains the list of manually assigned descriptors and a unique document identifier; followed by the full text immediately below (see **Figure 6**).

In order to allow users maximum usage flexibility and control, JEX processes are divided into **three main steps**: Pre-processing, indexing and post-processing. Users can develop their own pre-processing and post-processing modules to adapt the system to their own needs and to integrate it into their own environment, such as using their own visual interfaces, etc. Having the pre-processing step separate allows advanced users to experiment with the document representation (e.g. lemmatise input words, part-of-speech-tag the input document, segment the text into multi-word chunks, use word bigrams instead of single words, etc.). The post-processing step allows to change the format of the indexing result, or to add additional information to the results. For instance, the numerical descriptor code can be replaced by the descriptor name in a language of choice; the broader, narrower and related terms of each descriptor can be added to the output; the associates found in the text can be displayed together with each descriptor, etc.

**Fig. 5.** Output format of the main indexing phase.

**Fig. 6.** Compact format used as input to re-train JEX.

Users interested in training JEX will also be interested in **testing the indexing performance** of the newly trained classifiers. The software provides two ways of evaluating the system. The first method consists of an n-fold cross-validation performed on the whole training set. The second method allows to evaluate the performance of the system on a fixed set of documents, allowing an exact comparison between test runs.

Finally, the software release also includes a small **Java API** that allows users to access the different functionalities of the system directly from within other programs.

The released software has been readily trained for 22 languages on the document collection described in Table 1. For convenience, each language version is packaged together with the software. Users interested in indexing documents in more than one language will thus download several versions of the software.

The software runs on any modern computer without any specific specifications. The indexing process is extremely fast so that even large document collections can be indexed in little time, but training the system on document collections of the size described in Table 1 takes a few hours per language.

## 7. Acknowledgements

We would like to thank Bruno Pouliquen, who has developed a major part of the main assignment method, and Mladen Kolar, who has implemented an initial Java version of the tool. We would like to mention the support of Victoria Fernandez-Mera from the *Spanish Congress of Deputies* and Elisabet Lindkvist from the *Swedish Riksdagen*, who gave us a lot of advice on practices relating to manual EuroVoc indexing and who helped us to thoroughly evaluate the software. Finally, we are very grateful to the Publications Office of the European Commission for having provided their collection of manually EuroVoc-indexed documents. The initial work on JEX was funded as a JRC Exploratory Research project. The preparation of the first public release of JEX was funded under the JRC's Innovative Project Competition scheme.